\documentclass[letterpaper, 10 pt, conference]{ieeeconf}  

\IEEEoverridecommandlockouts                              

\usepackage{graphics} 
\usepackage{color}
\usepackage{comment}
\usepackage{graphicx}
\usepackage{wrapfig}
\usepackage{subcaption}
\usepackage{amsmath}
\usepackage{multirow}
\usepackage{hyperref}
\usepackage{url}
\usepackage{cleveref}
\usepackage{amssymb,amsmath} 

\title{\LARGE \bf
Subdimensional Expansion Using Attention-Based Learning For Multi-Agent Path Finding
}

\author{Lakshay Virmani$^{1*}$, Zhongqiang Ren$^{1*}$, Sivakumar Rathinam$^{2}$ and Howie Choset$^{1}$
	\thanks{$^{1}$ Lakshay Virmani, Zhongqiang Ren and Howie Choset are with Carnegie Mellon University, 5000 Forbes Ave., Pittsburgh, PA 15213, USA. 
	}%
	\thanks{$^{2}$Sivakumar Rathinam is with Texas A\&M University,
		College Station, TX 77843-3123.
	}
	\thanks{$^{*}$ The two authors contributed equally to this work.
	}%
}

\begin{document}

\maketitle
\thispagestyle{empty}
\pagestyle{empty}

\begin{abstract}
    Multi-Agent Path Finding (MAPF) finds conflict-free paths for multiple agents from their respective start to goal locations.
MAPF is challenging as the joint configuration space grows exponentially with respect to the number of agents.
Among MAPF planners, search-based methods, such as CBS and M*, effectively bypass the curse of dimensionality by employing a dynamically-coupled strategy: agents are planned in a fully decoupled manner at first, where potential conflicts between agents are ignored; and then agents either follow their individual plans or are coupled together for planning to resolve the conflicts between them.
In general, the number of conflicts to be resolved decides the run time of these planners and most of the existing work focuses on how to efficiently resolve these conflicts.
In this work, we take a different view and aim to reduce the number of conflicts (and thus improve the overall search efficiency) by improving each agent's individual plan.
By leveraging a Visual Transformer, we develop a learning-based single-agent planner, which plans for a single agent while paying attention to both the structure of the map and other agents with whom conflicts may happen.
We then develop a novel multi-agent planner called LM* by integrating this learning-based single-agent planner with M*.
Our results show that for both ``seen'' and ``unseen'' maps, in comparison with M*, LM* has fewer conflicts to be resolved and thus, runs faster and enjoys higher success rates.
We empirically show that MAPF solutions computed by LM* are near-optimal.
Our code is available at \url{https://github.com/lakshayvirmani/learning-assisted-mstar}.

\end{abstract}

\section{INTRODUCTION}\label{sec:intro}

Multi-Agent Path Finding (MAPF) aims to find collision-free paths for a team of agents from their respective start to goal locations while optimizing some path criteria, such as the sum of individual path lengths.
MAPF arises in many applications ranging from automated warehouses~\cite{wurman2008coordinating} to aircraft towing~\cite{morris2016planning}.
MAPF is challenging as the joint configuration space of agents grows exponentially with respect to the number of agents, and solving MAPF to optimality is NP-hard~\cite{10.5555/2891460.2891662}.

Among MAPF planners, search-based methods such as CBS~\cite{SHARON201540} and M*~\cite{WAGNER20151} effectively bypass this curse of dimensionality by employing a \emph{dynamically-coupled} strategy: At first, agents are planned in a fully \emph{decoupled} manner, where potential conflicts between agents are ignored; Then, agents either follow their individual plans or are coupled together for planning to resolve conflicts between them.
In general, the number of conflicts to be resolved between agents decides the run time of these dynamically-coupled MAPF planners and many of the existing work focus on how to efficiently resolve these conflicts~\cite{boyarski2015icbs,felner2018adding,6631119}.

\begin{figure}
      \centering
      \includegraphics[width=\columnwidth]{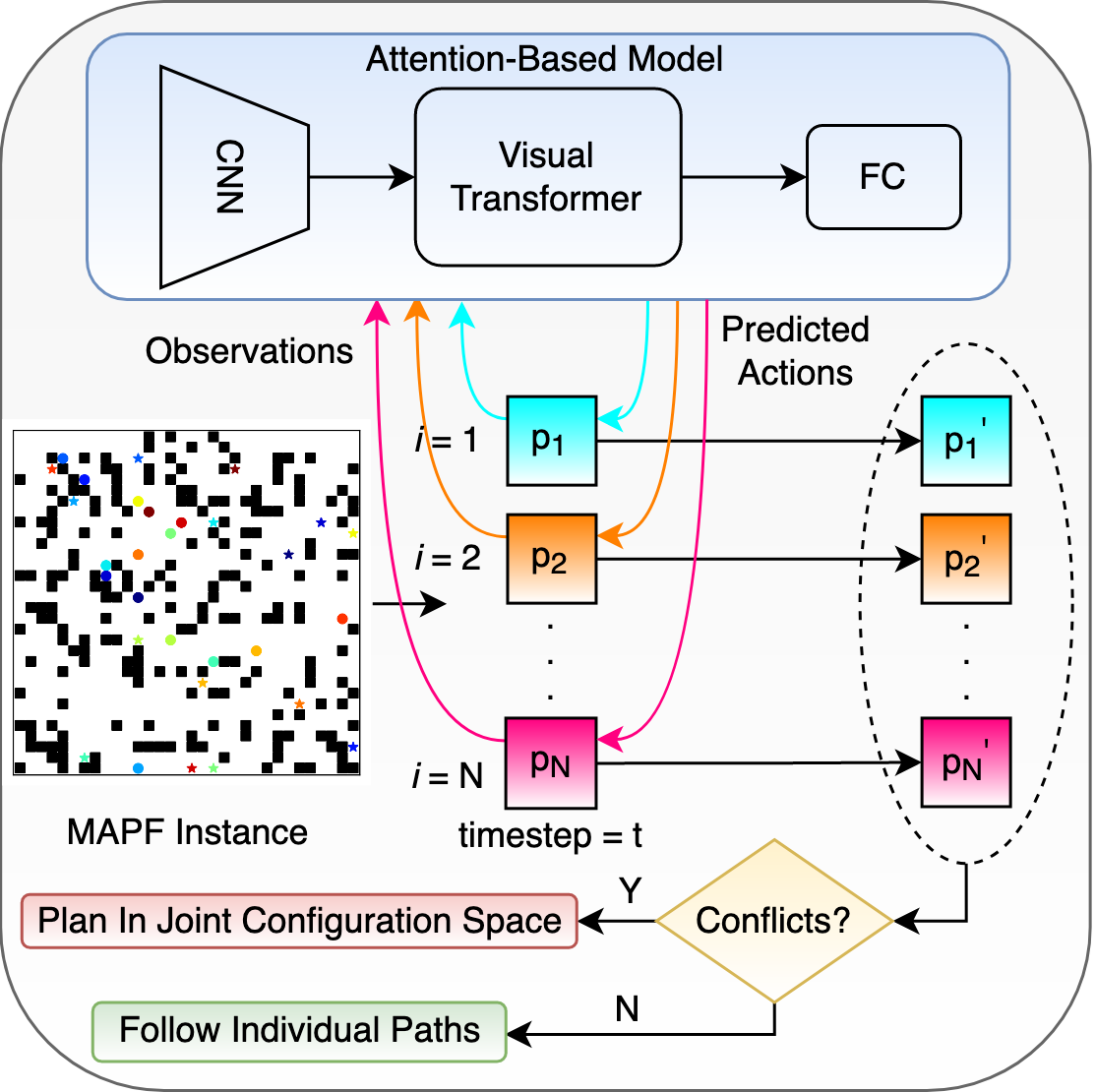}
      \caption{An illustration of Learning-Assisted M* (LM*).
      At every time step, each agent shares its observations with the attention-based model and the model predicts the action for each agent individually by attending to the structure of the map and other agents' information.
      The agents follow the predicted actions if there are no conflicts. Otherwise the agents in conflict are coupled together by planning in their joint configuration space, just like M*~\cite{WAGNER20151}.}
      \label{fig:overview}
      \vspace{-4mm}
\end{figure}

In this work, we take a different perspective and aim to reduce the number of potential conflicts between agents by improving their individual plans apriori in the first step.
Instead of planning for each agent in a fully decoupled manner (i.e. ignoring other agents), we take a learning-based approach to let each agent consider the potential conflicts with other agents (rather than ignoring them) and plan its individual path to avoid these potential conflicts (Fig.~\ref{fig:overview}).

Specifically, by leveraging a Visual Transformer~\cite{DBLP:journals/corr/abs-2006-03677}, we develop an \emph{attention}-based model that takes an agent's observation at each time step as input and output the (predicted) action of that agent.
This model can pay attention to the structure of the map and other agents' information (such as start, goals, etc.), in order to avoid potential conflicts.
The actions of an agent at all time steps computed by this model form an individual path from the start to the goal of that agent, and this model can thus be used as an individual planner for an agent.
With that in hand, we then develop a novel MAPF planner called LM* (Learning-assisted M*): LM* begins by using this attention-based model to plan an individual path for each agent, and then couples agents together when needed by planning in their joint configuration spaces (just as in M*) to resolve conflicts.

To verify the proposed LM*, we first leverage an expert MAPF planner ODrM*~\cite{6631119} to provide labeled data for training, and then test LM* against the original M* (i.e. baseline) in several different maps with various obstacle densities and different structures such as room, maze, etc.
Our results show that, (1) LM* can effectively reduce the number of conflicts among agents in comparison with M*, (2) LM* runs faster than M* in general (due to the reduced number of conflict), and (3) LM* empirically computes near-optimal solutions in a sense that the solution cost computed by LM* is less than 10\% more expensive than the one computed by M* (and M* is guaranteed to provide an optimal solution).

The rest of the article is organized as follows: Sec.~\ref{sec:prior} reviews related work and Sec.~\ref{sec:method} describes the method in detail.
We then discuss the training and the testing settings in Sec.~\ref{sec:experiment} along with the numerical results. We conclude and list our future work in Sec.~\ref{sec:conclusion}.

\section{Prior Works}\label{sec:prior}

\subsection{Multi-Agent Path Finding (MAPF)}

MAPF planners tend to fall on a spectrum from coupled~\cite{silver2005hca,10.5555/2898607.2898635} to decoupled~\cite{vcap2015complete}, trading off completeness and optimality for scalability.
In the middle of the spectrum, several planners take a dynamically-coupled strategy to efficiently compute conflict-free paths with optimality guarantees.
Among them, Conflict-Based Search (CBS) and its variants \cite{SHARON201540,Barer2014SuboptimalVO,boyarski2015icbs,ren21mocbs,andreychuk2019multi} employs a two-level search, where on the high level, conflicts between agents are detected and constraints on agents are generated, while on the low level, an individual path satisfying the added constraints for each agent is planned.

Subdimensional expansion \cite{WAGNER20151}, as another dynamically-coupled planner, begins by planning for each agent in a decoupled manner and then plans in the joint configuration space of agents to resolve conflicts.
Subdimensional expansion bypasses the curse of dimensionality by modifying the dimension of the search space based on agent-agent conflicts.
It inherits completeness and optimality if the underlying algorithm already has these features.
While being general to many planners \cite{wagner2012probabilistic,ren21momstar,ren21ms}, most of the prior work focus on applying subdimensional expansion to A*, which results in M*~\cite{WAGNER20151}.
In this work, we combine M* with attention-based learning~\cite{Vaswani2017AttentionIA,DBLP:journals/corr/abs-2006-03677} to avoid conflicts when planning individual paths for agents.

In addition to these dynamically-coupled MAPF planners, another set of related MAPF planners are the reinforcement learning (RL)-based methods~\cite{primal, mapper, DBLP:journals/corr/abs-2010-08184}.
These RL-based planners use multi-agent reinforcement and imitation learning to learn fully-decentralized ``end-to-end'' policies, which maps a partially observed world as well as other agents' information into actions for each agent for execution.
Different from RL-based planners that learn end-to-end policies, our LM* combines learning and search-based planner by ``embedding'' attention-based model into M* for the purpose of reducing the number of conflicts and improving the search efficiency of M*.

\subsection{Learning-Assisted Single-Agent Search}
Using learning techniques to improve single-agent path planning algorithms has been investigated recently~\cite{wang2020neural,ariki2019fully,Takahashi_Sun_Tian_Wang_2021,kaur2021speeding}.
For single-agent search-based planners, learning techniques are often leveraged to predict heuristic values~\cite{DBLP:journals/corr/BhardwajCS17,kaur2021speeding,Takahashi_Sun_Tian_Wang_2021,Li2020EfficientHG} of search states so that the search can be better guided towards the goal and the search effort can thus be reduced.
In these papers, the models are typically trained to predict the heuristic value of all possible states within the search space in order to guide the search.

However, all these methods are limited to a single-agent.
Applying them directly to a multi-agent system is non-trivial for the following two reasons.
On one hand, the interactions between agents are not considered in the single-agent model, which is important for multi-agent problems.
On the other hand, if we treat all agents as a single ``meta-agent'' and apply these single-agent methods to the corresponding joint configuration space, the curse of dimensionality makes these methods scale poorly.
To handle this challenge, by leveraging the self-attention mechanism~\cite{Vaswani2017AttentionIA}, we choose to train a single-agent model that can predict the next move for an agent while paying attention to the structure of the map and other agents with whom interactions may happen.

\subsection{Attentions and Transformers}
Transformers \cite{Vaswani2017AttentionIA} were first
introduced as a new attention-based method for machine translation.
They introduced self-attention \cite{Bahdanau2015NeuralMT} layers which scan through each element of a sequence and aggregate information from the whole sequence.
They are now widely used in various applications~\cite{DBLP:journals/corr/abs-1905-03072,DBLP:journals/corr/abs-1708-09545}.
Transformers, and more specifically, the self-attention mechanisms are able to model the \emph{relations} among per-elements such as tokens (for sequence data \cite{Vaswani2017AttentionIA,DBLP:journals/corr/abs-1810-04805}) and pixels (for image data \cite{DBLP:journals/corr/abs-2005-12872}).
The capability to model relations inspires us to leverage transformers to describe interactions (i.e. collision avoidance) between agents for MAPF.

Specifically, we take the view that transformers can help each agent to pay attention to the structure of the map and the subset of other agents that they have to interact with in order to avoid conflicts.
However, applying transformer to a multi-agent system is challenging due to the curse of dimensionality of multi-agent systems as well as the computational burden of self-attention operations.
To circumvent this challenge, we leverage the recent Visual Transformer~\cite{DBLP:journals/corr/abs-2006-03677} which operates in a semantic token space, judiciously attending to parts of the input based on the ``context''.
Using token space helps preserve the most important features and considerably reduces the number of parameters for the self-attention operation in Transformers.

\section{Our Approach}\label{sec:method}

\subsection{Problem Description}

Let index set $I = \{1,2,\dots,N\}$ denote a set of $N$ agents.
All agents move in a workspace represented as a finite graph $G=(V,E)$, where the vertex set $V$ represents the possible locations for agents and the edge set $E =V \times V$ denotes the set of all the possible actions that can move an agent between any two vertices in $V$.
An edge between $u,v\in V$ is denoted as $(u, v)\in E$ and the cost of an edge $e \in E$ is a non-negative real number cost$(e) \in \mathbb{R}^{+}$.

Let $\pi^i(v^i_{1}, v^i_{\ell})$ denote a path for agent $i$ that connects vertices $v^i_{1}$ and $v^i_{\ell}$ via a sequence of vertices $(v^i_{1},v^i_{{2}},\dots,v^i_{\ell})$ in the graph $G$. 
Let $g^i(\pi^i(v^i_{1}, v^i_\ell))$ denote the cost associated with the path.
This path cost is the sum of the costs of all the edges present in the path, $i.e.$, $g^i(\pi^i(v^i_{1}, v^i_{\ell})) = \Sigma_{j=1,2,\dots,{\ell-1}} cost(v^i_{{j}}, v^i_{{j+1}})$. 

All agents share a global clock. Each action, either wait or move, requires one unit of time for any agent.
Any two agents $i,j \in I$ are claimed to be in conflict if one of the following two cases happens.
The first case is a ``vertex conflict'' where two agents occupy the same vertex at the same time.
The second case is an ``edge conflict'' (also called swap conflict) where two agents travel through the same edge from opposite directions between times $t$ and $t+1$ for some $t$.

Let $v_o^i,v^i_d\in V$ denote the start and goal vertex of agent $i$. The Multi-Agent Path Finding (MAPF) problem aims to compute conflict-free paths for all agents while the sum of path costs reaches the minimum.

\subsection{Approach Overview}
As shown in Fig.~\ref{fig:overview}, we develop an attention-based model that is able to ``plan path'' for a \emph{single} agent while taking other agents and the map into consideration.
Specifically, the model takes $O^i_t$, the observation of agent $i$ at time step $t$ (see Sec.~\ref{inputs}) as input, outputs the next action $A^i_{t}$ of that agent, and all of the computed actions $A^i_{t},t=0,1,\dots,T$, where $T$ is the time step where all agents have arrived at their respective goals, forms an individual path $\pi^i$ for agent $i$.
During the training phase, the model uses data generated by solving various MAPF instances using an expert MAPF planner ODrM*~\cite{6631119}. 
During the testing phase, each agent shares its observation $O^i_t$ as input to the model and the model predicts the action $A^i_t$ for the agent.

The model begins with a number of convolutional layers to extract the low-level features from $O^i_t$.
The resulting output feature map then passes through a Visual Transformer (VT).
The VT first uses a tokenizer to group pixels (of the feature map) into a small number of visual tokens; with each token representing a semantic concept in the image, 
Transformers are then applied to model relationships between these tokens.
The attended visual tokens are then used as input to fully connected layers, which then output the predicted actions for agents.

\subsection{Learning-Assisted M*}
In this section, we describe how our attention-based model is integrated with M*.
The regular M*~\cite{WAGNER20151} begins by running exhaustive backwards A* searches to compute an individual optimal policy $\phi^i$ for each agent $i$, where $\phi^i$ takes the current location of the agent and returns the next optimal move.
The search process of M* is guided by these policies in a sense that agents either follow their individual policies if there is no conflict or are coupled together by planning in the joint configuration space to resolve conflicts.

In this work, instead of running a backwards A* search for each agent (which ignores any other agents), we compute these policies using the aforementioned attention-based model, which takes each agent's observations and returns a next move for that agent.
Since the output of both $\phi^i$ and the attention-based model are the same (i.e. the next move of an agent), the model can be readily integrated into M*.
We name our approach \emph{learning-assisted M* (LM*)} since learning-based method (i.e. the attention-based model) is leveraged to learn to predict and thus avoid conflicts when planning individual paths for each agent.
For the rest of this section, we present this attention-based model in details.

\subsection{Action Space}
In this work, we consider the graph $G$ to be a four-connected grid in which agents are allowed to move in one of the four cardinal directions or to wait in place at each time step.
Each action, either wait or move takes a unit time and incurs a unit cost.
Moving into an obstacle is considered to be an invalid move, and if an agent selects to move into an obstacle during testing, it instead waits in place for that time step. 
In practice, after training, agents rarely choose an invalid move, which indicates that they effectively learn the set of valid actions at each location.

\subsection{Observation Space} \label{inputs}
The observation $O^i_t$ of an agent consists of ten channels, where each channel comprises of a 32x32 size matrix.

\subsubsection{Obstacles}
Channel 1 contains a binary matrix which represents the free space and obstacles in the grid graph. Specifically, entries corresponding to free spaces have value zeros while entries of obstacles have value ones.

\subsubsection{Agent $i$ Start, Goal, and Cost-to-go}
Channel 2 and 3 consists of two matrices describing agent $i$'s start and goal respectively.
In these matrices, the entry corresponding to the agent $i$'s start (or goal) has value one while all other entries have value zeros. We then generate another matrix (Channel 4) in which each entry has a value of the minimum cost-to-go from the corresponding location to agent $i$'s goal $v^i_d$.
These values are computed by running Dijkstra search backwards from $v^i_d$ while ignoring any other agents.
We scale this matrix so that all values lie between 0 and 1 (i.e. normalize).

\subsubsection{Other Agents’ Starts, Goals, and Sum of Cost-to-go}
Similarly, channel 5 and 6 of $O^i_t$ are two binary matrices that represent the start and goal locations of all other agents $(j \in \{1,2,...,N\}, j \neq i)$ respectively, where the entries corresponding to any other agents' starts (or goals) have value ones while all other entries have value zeros.
We also compute the cost-to-go for each agent $j$, sum up these cost values for each location and normalize, which results in a matrix of channel 7.
Intuitively speaking, these 3 matrices (i.e. channel 5, 6, 7) together provide the model the context about where the other agents in the world are headed and what route they might take.

\subsubsection{Other Agents’ Future Positions}
As shown in PRIMAL2 \cite{DBLP:journals/corr/abs-2010-08184}, an RL-based MAPF planner, considering the future positions of other agents can help avoid conflicts.
In this work, similar to PRIMAL2 \cite{DBLP:journals/corr/abs-2010-08184}, we use three binary matrices to provide the future position of other agents, one per time step.
These matrices constitute channel 8, 9 and 10 in $O^i_t$.
It's worthwhile to point that, in our experiments, we observed that adding channels 6-10 in $O^i_t$ can help the model to scale to a larger number of agents, and the model achieves higher train and test accuracy.

\subsection{Model Architecture}
As shown in Figure \ref{fig:overview}, our model comprises of 3 main components: (1) a convolutional neural network (CNN) to learn densely-distributed, low-level patterns, (2) followed by a Visual Transformer to learn and relate more sparsely-distributed, higher-order semantic concepts, and (3) finally fully connected layers for action classification.

\begin{figure}
      \centering
      \includegraphics[width=0.55\linewidth]{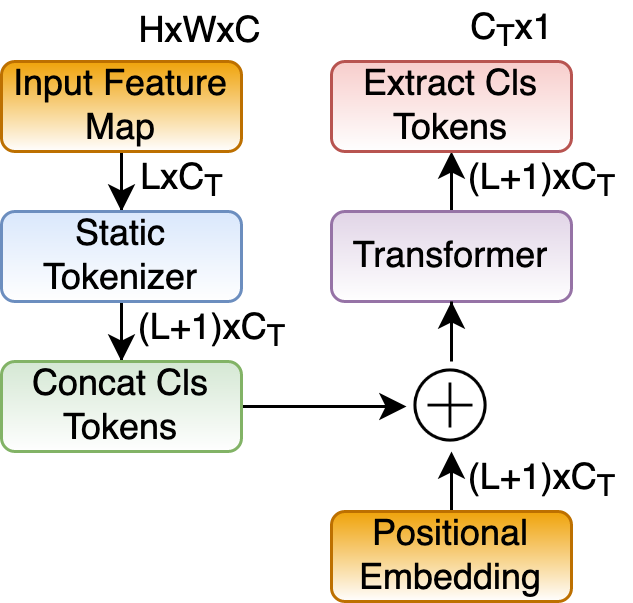}
      \caption{Visual Transformer comprises of a static tokenizer and a transformer. Classification (Cls) tokens and positional embeddings are used to enable attention-based learning within the transformer.}
      \label{fig:vt}
      \vspace{-5mm}
\end{figure}

We use 3 residual learning building blocks (BasicBlock) inherited from ResNet \cite{DBLP:journals/corr/HeZRS15} as the CNN backbone of our model (Fig.~\ref{fig:detailed_arch}). The output feature map from these convolutions is then passed through a Visual Transformer (VT) (Fig.~\ref{fig:vt}). VT uses a static tokenizer module to convert the feature map into a compact set of visual tokens. Formally, a feature map can be represented by \(X  \in \mathbf{R}^{H \times W \times C}\), where \(H,W\) are the height and width of the feature map, and \(C\) is the channel size of the feature map. Consider \(\bar{X} \in \mathbf{R}^{HW \times C}\) to be a reshaped matrix of X obtained by merging the two spatial dimensions into one. Visual tokens can be represented by \(T \in \mathbf{R}^{L \times C_T}\) where \(L\) is the number of visual tokens, and \(C_T\) is the channel size of the visual tokens.
The static tokenization can be described by:
\vspace{-1mm}
\begin{eqnarray}
A &=& softmax(\bar{X}W_A^T) \\
V &=& \bar{X}W_V \\
T &=& A^TV = \sum_i A[i,:]^T V[i,:].
\vspace{-2mm}
\end{eqnarray}

Here, \(A \in \mathbf{R}^{HW \times L}\) are normalized token coefficients, and the value of \(A[i; l] \in \mathbf{R}\) determines the contribution of the \(i\)-th pixel \(V [i; :]\) to the \(l\)-th token \(T[l; :]\). \(W_A \in \mathbf{R}^{L \times C}\) and \(W_V \in \mathbf{R}^{C \times C_T}\) are learnable weights used to compute \(A\) and to convert the feature map \(X\) into \(V \in \mathbf{R}^{HW \times C_T}\) respectively.

In order to perform classification, we use the approach \cite{DBLP:journals/corr/abs-1810-04805} of adding an extra learnable “classification token” to the visual tokens extracted from the feature map. Position embeddings are also added to retain positional information as done in \cite{DBLP:journals/corr/abs-2010-11929}.
The resulting vectors are then fed to the transformer \cite{DBLP:journals/corr/abs-2010-11929}, which can be described as:
\begin{equation}
T_{out} = T_{in} + (softmax((T_{in}K)(T_{in}Q)^{T})(T_{in}V))F,
\end{equation}
where \(T_{in}, T_{out} \in \mathbf{R}^{(L+1) \times C_T}\) are the input and output tokens. \(K \in \mathbf{R}^{C_T \times C_T/2}, Q \in \mathbf{R}^{C_T \times C_T/2}, V \in \mathbf{R}^{C_T \times C_T}, F  \in \mathbf{R}^{C_T \times C_T}\) are learnable weights used to compute keys, queries, values, and output tokens.
We then use the attended classification tokens and cross-entropy loss to train the model for action classification.

\section{Experiments}\label{sec:experiment}

\newcommand{\mydata}[3]{{\small#1 / {\color{red} #2} / {\color{blue} #3}}}

\subsection{Implementation Detail}

\begin{figure}[tb]
      \centering
      \includegraphics[width=0.9\columnwidth, height=46mm]{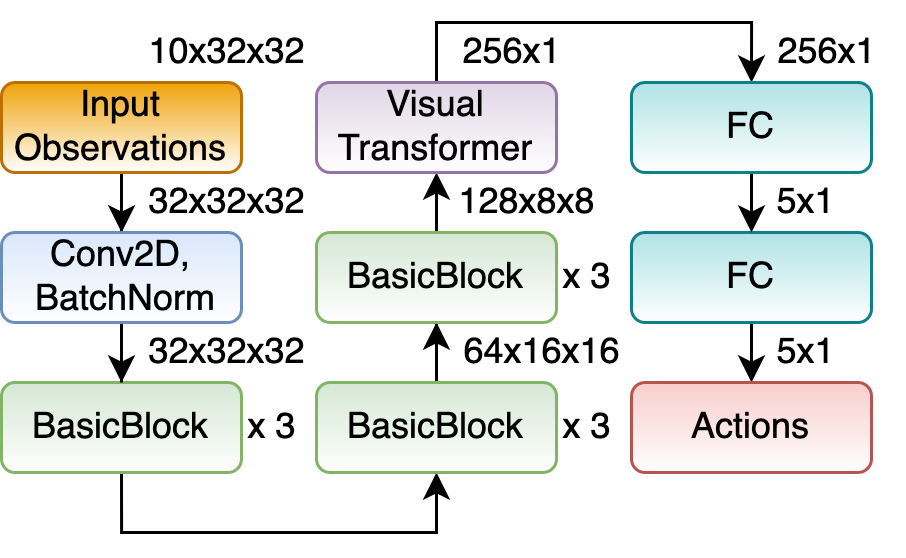}
      \caption{The network consists of multiple convolutional layers followed by a Visual Transformer and 2 fully connected layers.
      The dimensions to the top right corner of the boxes represent the output sizes of the module.}
      \label{fig:detailed_arch}
      \vspace{-3mm}
\end{figure}

As shown in Fig.~\ref{fig:detailed_arch}, we stack $3$ BasicBlock \cite{DBLP:journals/corr/HeZRS15} with an output channel size of $32$, $64$, and $128$ respectively, followed by a Visual Transformer (VT).
The input to the VT is a feature map of size $H = 8$, $W = 8$ and $C = 128$.
We extract $L = 16$ tokens from the feature map with a channel size of $C_T = 256$.
The visual tokens are attended to using a transformer comprising of 16 encoder layers each containing multi-headed attention modules with 16 attention heads and multi-layer perceptron with a dimension of $512$.

We implement the model in Python using the Pytorch \cite{NEURIPS2019_9015} library.
The model is trained and tested with a 3.30 GHz Intel(R) Core(TM) i9-9820X CPU and a NVIDIA GeForce RTX 2080 Ti GPU.
During the training, we minimize the cross-entropy loss using the Adam optimizer \cite{DBLP:journals/corr/KingmaB14} for 10 epochs.
We use batch size of 64 and start with a learning rate of 0.003 which is decayed by a factor of 0.992 after every 10k steps.
We obtain a top-1\% accuracy of 91.3\% on the training set and 91.7\% on the testing set.
Such high train and validation accuracy indicates that the model is capable to select an optimal action for each agent during the planning process.
To compare M* and LM* during the test, both algorithms are given 300 seconds to solve each MAPF instance.
We test with three different inflation rates: 1.0, 1.1 and 10.0, which are used in M*~\cite{WAGNER20151} and PRIMAL~\cite{primal}.

\subsection{Data Generation}\label{sec:data_gen}

\begin{figure}[tb]
      \centering
      \hspace*{-0.5cm}  
      \includegraphics[width=0.8\columnwidth]{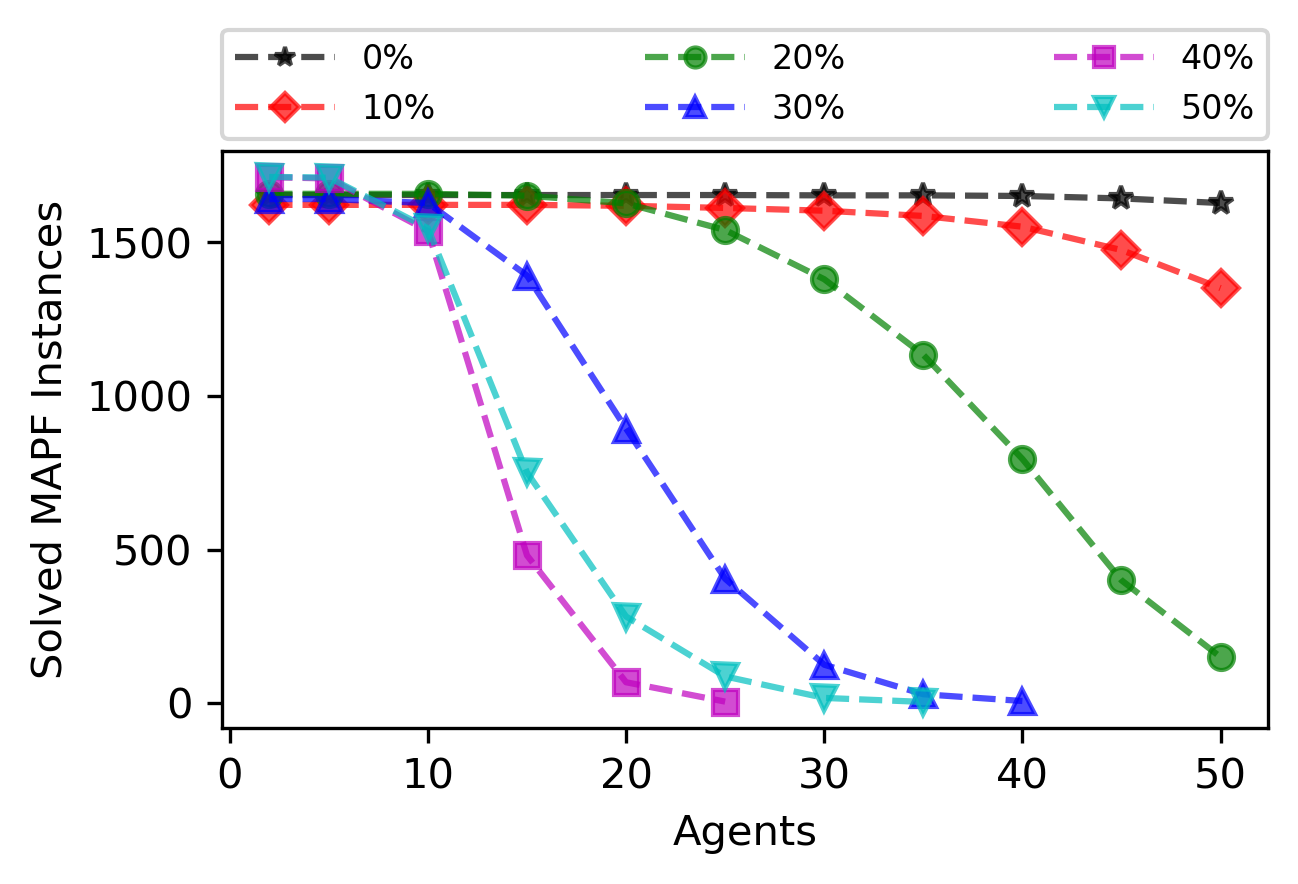}
      \vspace{-2mm}
      \caption{The number of solved MAPF instances in the training set with respect to the number of agents, under different obstacle probability.}
      \vspace{-4mm}
      \label{fig:datasetdist}
\end{figure}

For training, we generate 10k (k stands for thousand) random maps.
Each map has a size of 32x32 and the obstacles are placed randomly, where the probability of each cell being marked as an obstacle is randomly
selected from $\{0\%, 10\%, 20\%, 30\%, 40\%, 50\%\}$.
In each map, we generate a MAPF instance with $N=2,3,\dots,50$ respectively, where $N$ is the number of agents in that instance.
Thus, we generate 490k MAPF instances, 10k for each number of agents from 2 to 50.
For each test instance, a unique start and goal is selected randomly for each agent, and it is ensured that a path from the start to the goal exists.

To provide labeled training data $\{(O^i_t, A^i_t)\}$ (i.e. the observation of an agent at a certain time step and an action that should be selected by the agent at that time step), we use ODrM* \cite{6631119} with an inflation of 1.1 and a time limit of 300 seconds to solve these MAPF instances.
Out of the 490k MAPF instances, ODrM* is able to solve approximately 295k instances.
Fig. \ref{fig:datasetdist} shows the number of solved instances with respect to the number of agents, which characterize the distribution of the training data set.
For each solved MAPF instance, the solution is a joint path $\pi=\{v_1,v_2,\dots,v_\ell\}$, where each $v_k=(v^1_k,v^2_k,\dots, v^N_k), k=1,2,\dots,\ell$ is a joint vertex that contains the locations of all agents. To generate training data $\{(O^i_t, A^i_t)\}$, we first select 30\% of the joint vertices $v_k$ from the joint path $\pi$.
We then further select 30\% of the agents and their corresponding individual vertices $v^i_k$ from each joint vertex $v_k$ and generate labels using the observation $O^i_t$ of the agent at that time step and the action $A^i_t$ the agent takes in order to move to $v^i_{k+1}$.
Using this process we are able to get a dataset of size 23 million, which is then split into the train set (90\%) and test set (10\%).

\subsection{Seen Maps, Unseen Agent Positions}

\begin{figure}[tb]
      \centering
      \includegraphics[width=\columnwidth]{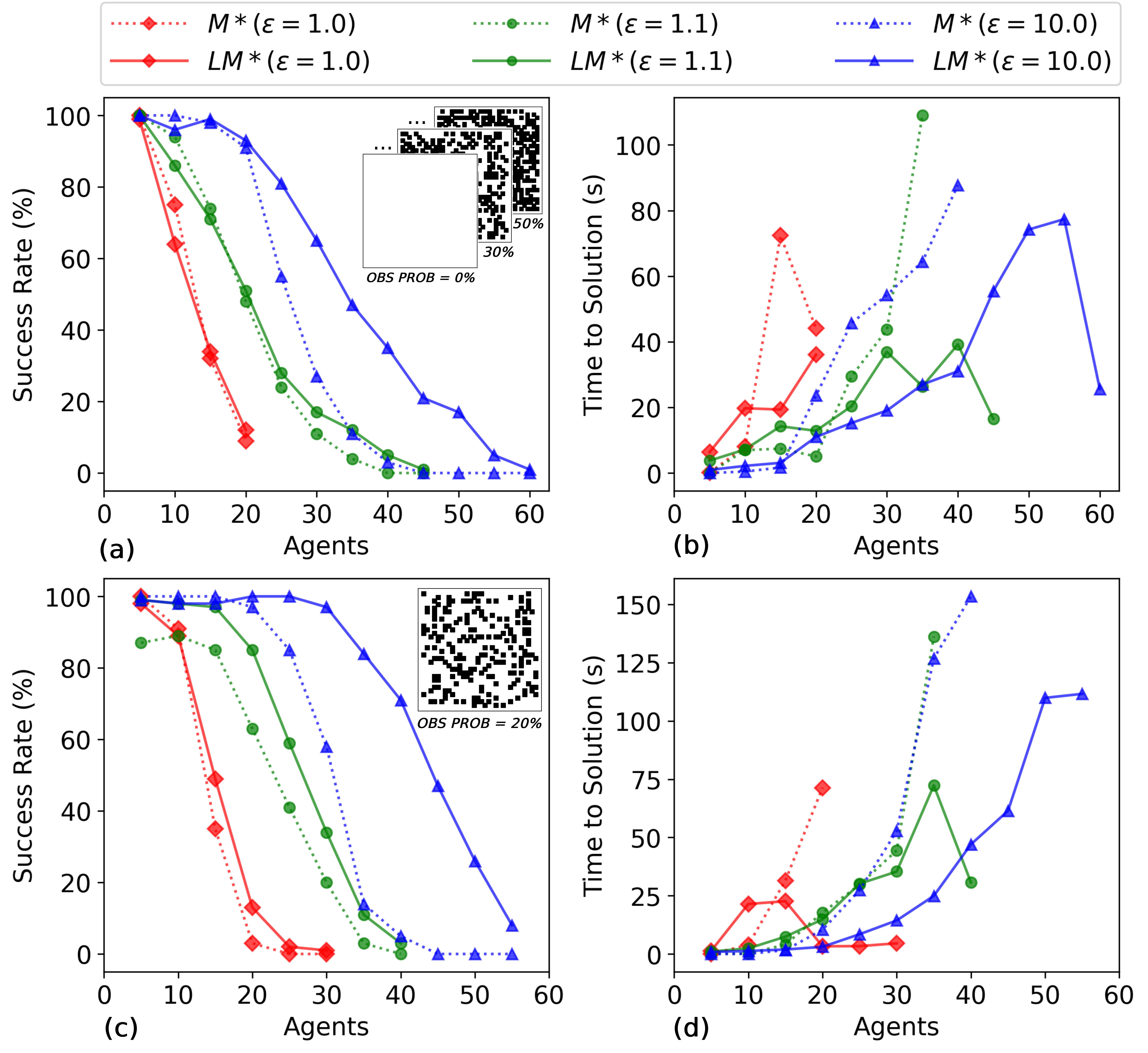}
      \caption{Success rates and the average run time to solution for different maps. ``OBS PROB'' stands for the obstacle probability of the map.}
      \label{fig:sr_t}
      \vspace{-4mm}
\end{figure}

To compare the performance of M* and LM*, we first start to test on maps that appear in the training set, and hence the name ``seen maps'', as LM* has ``seen'' these maps during the training.
We randomly choose 100 maps from the train set and generate test instances with $N=5,10,15...60$ agents (similar to Sec. \ref{sec:data_gen}).
The success rates and the average run time to solution are shown in Fig. \ref{fig:sr_t} (a), (b).
LM* enjoys higher success rates than M* and shorter average run times in general, within the seen maps.

\begin{table}[htb]
\begin{center}
\tabcolsep=0.07cm
\renewcommand{\arraystretch}{1.13}
\begin{tabular}{|c|c|c|c|}
\hline
\multirow{2}{*}{Agents} & \multicolumn{3}{c}{\mydata{$\epsilon = 1.0$}{$\epsilon = 1.1$}{$\epsilon = 10.0$}} \vline \\
& Max Collision Set Size & Nodes Generated & Nodes Expanded\\
\hline
5 & \mydata{45}{42}{42} & \mydata{71}{10}{16} & \mydata{78}{-1}{1}\\
\hline
10 & \mydata{43}{33}{26} & \mydata{69}{80}{90} & \mydata{91}{85}{76}\\
\hline
15 & \mydata{66}{37}{30} & \mydata{95}{66}{70} & \mydata{96}{53}{75}\\
\hline
20 & \mydata{41}{17}{20} & \mydata{86}{-1}{83} & \mydata{78}{-21}{31}\\
\hline
25 & \mydata{-}{15}{17} & \mydata{-}{53}{84} & \mydata{-}{25}{70}\\
\hline
30 & \mydata{-}{8}{6} & \mydata{-}{54}{79} & \mydata{-}{2}{51}\\
\hline
35 & \mydata{-}{-}{8} & \mydata{-}{-}{74} & \mydata{-}{-}{13}\\
\hline
\end{tabular}
\end{center}
\caption{This tables shows the percentage of decreases offered by LM* in comparison with M* in ``seen'' maps against the metrics: (left) the maximum size of the collision set during the planning, (middle) the number of nodes generated, and (right) the number of nodes expanded.}
\label{tab:seen}
\vspace{-2mm}
\end{table}

Table~\ref{tab:seen} explains the reason for such improvements. As shown in the table, the largest size of the collision set \cite{WAGNER20151}, which can be regarded as a metric in M* to describe the number of conflicts between agents, is reduced up to 66\% when $N=15,\epsilon=1.0$. Consequently, the number of nodes being generated and expanded is also reduced (up to 95\%). It shows that in seen maps, LM* is able to circumvent conflicts between agents and thus enhances the success rates and reduces the average run time.
Finally, as shown in Table~\ref{tab:cost_comp}, for most ($>90\%$) of the test instances, the solution cost computed by LM* is less than 10\% more expensive than the solution cost computed by M*.
It shows that, empirically, LM* is able to produce near-optimal solutions.

\subsection{Unseen Maps, Same Type}

This section verifies whether LM* is able to generalize to unseen maps that are of the same type (i.e. random) as the maps in the training set.
We generate the test instances using the convention as one in M*~\cite{WAGNER20151}: the grid map is of size 32x32 and each cell has a 20\% probability to be occupied by an obstacle.
Unique start and goal locations for each agent are chosen randomly and the existence of a path connecting the start and the goal is ensured.

\begin{table}[h]
\begin{center}
\tabcolsep=0.07cm
\renewcommand{\arraystretch}{1.13}
\begin{tabular}{|c|c|c|c|}
\hline
\multirow{2}{*}{Agents} & \multicolumn{3}{c}{\mydata{$\epsilon = 1.0$}{$\epsilon = 1.1$}{$\epsilon = 10.0$}} \vline \\
& Max Collision Set Size & Nodes Generated & Nodes Expanded\\
\hline
5 & \mydata{76}{74}{73} & \mydata{82}{27}{49} & \mydata{71}{10}{8}\\
\hline
10 & \mydata{52}{45}{52} & \mydata{49}{-47}{90} & \mydata{67}{-4}{17}\\
\hline
15 & \mydata{57}{34}{45} & \mydata{85}{38}{99} & \mydata{74}{8}{34}\\
\hline
20 & \mydata{94}{25}{37} & \mydata{99}{59}{99} & \mydata{92}{51}{49}\\
\hline
25 & \mydata{-}{19}{29} & \mydata{-}{28}{89} & \mydata{-}{16}{43}\\
\hline
30 & \mydata{-}{17}{17} & \mydata{-}{45}{87} & \mydata{-}{52}{50}\\
\hline
35 & \mydata{-}{-}{12} & \mydata{-}{-}{88} & \mydata{-}{-}{52}\\
\hline
\end{tabular}
\end{center}
\caption{This tables shows the percentage of decreases offered by LM* in comparison with M* in ``unseen'' maps against the same metrics as in Table \ref{tab:seen}.}
\label{tab:unseen}
\vspace{-3mm}
\end{table}

Fig. \ref{fig:sr_t} (c), (d) show the success rates and the average run time to solution for different numbers of agents.
LM* achieves higher success rates and lower run time than M* in general, which indicates the generalization capability of LM* to unseen maps of the same type.

\begin{figure}[bth]
\vspace{-3mm}
      \centering
      \includegraphics[width=0.6\columnwidth]{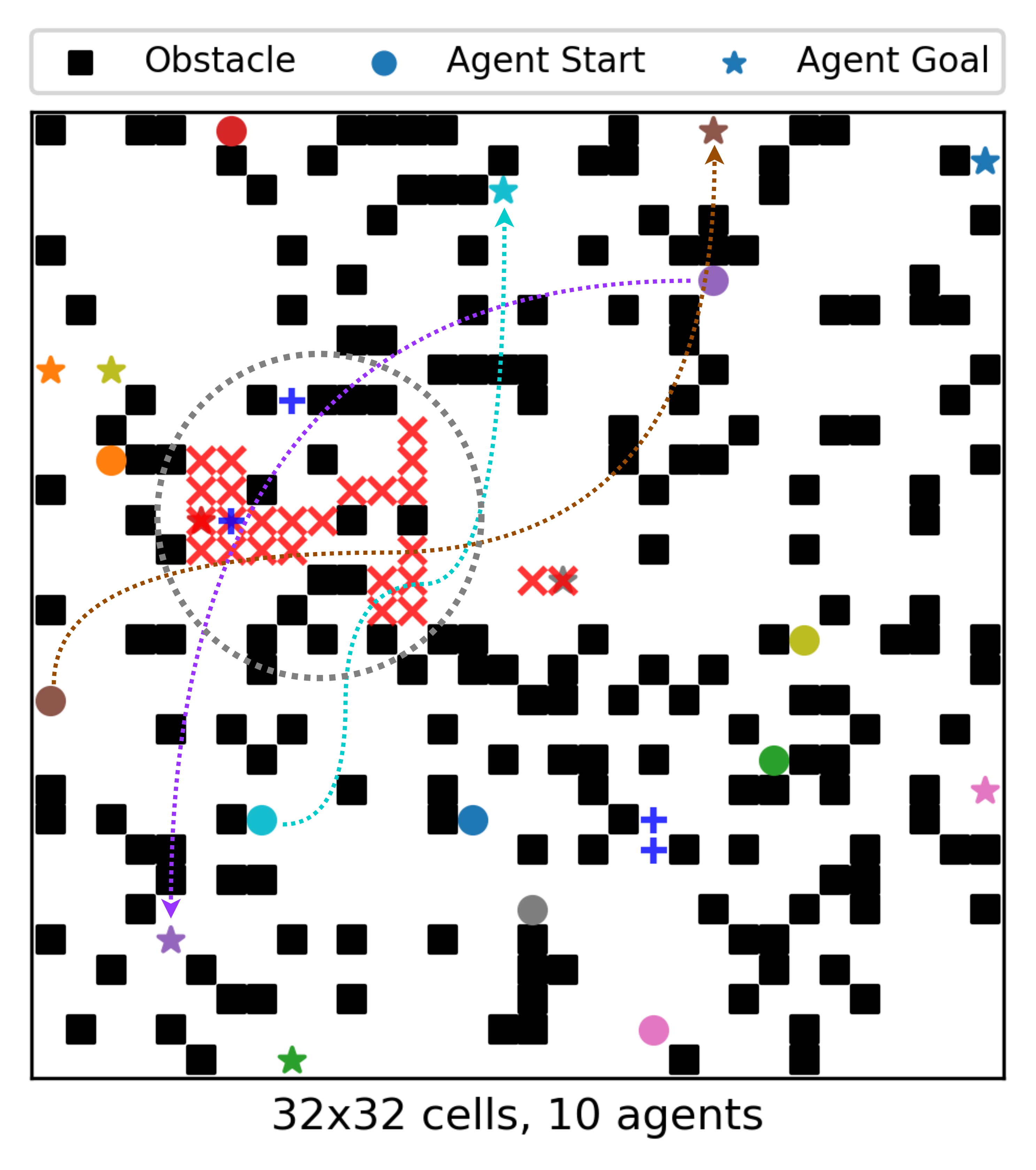}
      \caption{A sample MAPF test instance, where the red ``x'' and the blue ``+'' mark the locations where agent-agent conflicts are detected during the search in M* and LM* respectively.}
      \label{fig:practical}
      \vspace{-3mm}
\end{figure}

Furthermore, Fig. \ref{fig:practical} shows a sample instance from the test set.
The red ``x'' shows the locations where conflicts between agents are detected and M* has to resolve these conflicts during the planning.
The blue ``+'' shows the locations where conflicts between agents are detected in LM*, which is notably much fewer than the number of conflicts in M*.
As a result, LM* achieves higher success rates and shorter run time.

\begin{table}[tbh]
\begin{center}
\tabcolsep=0.07cm
\renewcommand{\arraystretch}{1.13}
\begin{tabular}{|c|c|c|}
    \hline
    \multirow{2}{*}{Test Type} & \multicolumn{2}{c}{\mydata{$\epsilon = 1.0$}{$\epsilon = 1.1$}{$\epsilon = 10.0$}} \vline \\
    & \% Instances & Max. \% Cost Incre.\\
    \hline
    Seen & \mydata{98.3}{99.1}{91.4} & \mydata{30.2}{30.3}{97.2} \\
    \hline
    Unseen, Same & \mydata{100.0}{99.7}{95.8} & \mydata{2.0}{15.9}{28.0} \\
    \hline
\end{tabular}
\end{center}
\caption{The left column shows the percentage of instances solved by LM* in which the solution cost is less than 10\% more expensive than the solution cost computed by M*. The right column shows the maximum (i.e. worst-case) ratio between the solution cost computed by LM* and the solution cost computed by M*.}
\label{tab:cost_comp}
\vspace{-4mm}
\end{table}

\subsection{Unseen Maps, Different Types}

Finally, we verify whether LM* can generalize to unseen map of different types (such as room-like grid) as the training set (random grid).
Here, we use the room (room-32-32-4) and maze maps ('maze-32-32-2', 'maze-32-32-4') from a online data set~\cite{stern2019mapf}.
Due to the space limit, we omit the plots and summarize the results:
In the room map, LM* achieves better success rates and shorter run time than M* in general, but in the maze-like maps, LM* fails to solve any instances when $N>10$, while M* can still solve some of the instances.
The possible reason is that: room-like map is ``similar'' to the random maps that appears in the training set while maze-like maps are drastically different as compared to the random maps in the training set.
This result leads us to further explore how to improve LM* so that it can handle unseen maps with very different structures, in our future work.

\section{CONCLUSION}\label{sec:conclusion}

In this work, we introduce a novel MAPF planner called Learning-assisted M* (LM*) by leveraging both attention-based learning~\cite{DBLP:journals/corr/abs-2006-03677} and M*~\cite{WAGNER20151}.
LM* begins by running an attention-based model to plan for each agent, and couples agents together to plan in their joint configuration space to resolve conflicts between agents.
Our results show that LM* is able to circumvent conflicts between agents and thus achieves higher success rates and shorter run time.

For future work, one can investigate whether the developed attention-based model can be fused with other MAPF planners such as CBS~\cite{SHARON201540}.
Additionally, one can also consider further improve this attention-based model to handle environments that are very different to the ones appeared in the training set.







\section*{ACKNOWLEDGMENT}

This material is based upon work supported by the National Science Foundation under Grant No. 2120219 and 2120529. Any opinions, findings, and conclusions or recommendations expressed in this material are those of the author(s) and do not necessarily reflect the views of the National Science Foundation.

\bibliographystyle{plain}
\bibliography{ref}

\end{document}